%% file: main.tex
\documentclass[runningheads]{llncs}
\usepackage[T1]{fontenc}
\usepackage{graphicx}
\usepackage{subcaption}
\usepackage{graphicx}
\usepackage{color}
\usepackage{url}

\usepackage{todonotes}

\usepackage{epsfig}
\usepackage{calc}
\usepackage{amssymb}
\usepackage{amstext}
\usepackage{amsmath}
\usepackage{multicol}
\usepackage{pslatex}
\usepackage{algorithm2e}
\usepackage[bottom]{footmisc}
\usepackage{physics}
\usepackage{tikz}
\usetikzlibrary{quantikz2}
\usepackage[noend]{algpseudocode}
\usepackage{pgfplots}
\usepackage[outline]{contour}
\contourlength{1.2pt}
\usepackage{placeins}
\pgfplotsset{compat=1.18}

\usepackage[hidelinks]{hyperref}
\usepackage[capitalise]{cleveref}
\creflabelformat{equation}{#2#1#3}

\begin{document}
\title{Efficient Quantum One-Class Support Vector Machines for Anomaly Detection Using Randomized Measurements and Variable Subsampling} 
\titlerunning{Quantum OC-SVMs Using Randomized Measurements and Variable Subsampling}
%
\author{
Michael Kölle\inst{1}* \and 
Afrae Ahouzi\inst{1}* \and 
Pascal Debus\inst{2} \and 
Elif Çetiner \inst{1} \and 
Robert Müller\inst{1} \and 
Daniëlle Schuman\inst{1} \and 
Claudia Linnhoff-Popien\inst{1}}
\authorrunning{M. Kölle et al.}
%
\institute{LMU Munich, Oettingenstraße 67, 80538 Munich, Germany \and
Fraunhofer AISEC, Garching, Germany\\
\email{michael.koelle@ifi.lmu.de}}

\maketitle 

\begin{abstract}
\input{content/0_abstract}
\end{abstract}

\def\thefootnote{*}\footnotetext{Combined first authorship.}
\input{content/1_introduction}
\input{content/2_preliminaries}
\input{content/3_related_work}

\input{content/4_approach}

\input{content/5_experimental_setup}

\input{content/6_results}
\input{content/7_conclusion}
\input{content/8_acknowledgements}

%
\bibliographystyle{splncs04}
\bibliography{main}
\end{document}

%% file: content/0_abstract.tex
Quantum one-class support vector machines leverage the advantage of quantum kernel methods for semi-supervised anomaly detection. However, their quadratic time complexity with respect to data size poses challenges when dealing with large datasets. In recent work, quantum randomized measurements kernels and variable subsampling were proposed, as two independent methods to address this problem. The former achieves higher average precision, but suffers from variance, while the latter achieves linear complexity to data size and has lower variance. The current work focuses instead on combining these two methods, along with rotated feature bagging, to achieve linear time complexity both to data size and to number of features. Despite their instability, the resulting models exhibit considerably higher performance and faster training and testing times. 

%% file: content/1_introduction.tex
\section{Introduction} \label{sec:introduction}

Anomaly detection algorithms ensure the proper functioning and security of various systems in today's increasingly digitalized landscape. They aim to identify observations or events that deviate from expected patterns within a dataset.
They can detect irregularities such as unauthorized network access or unexpected machinery behavior, thus preventing potential data breaches, accidents, and financial losses. In the medical field, these algorithms enable earlier and more accurate diagnoses \cite{fernando2021deep}, reduce the incidence of medical errors, and facilitate real-time monitoring of patients' conditions by alerting healthcare providers to sudden changes \cite{9377017}. Similarly, in electronic commerce, they are essential for enabling authorized customers to transact securely online while protecting financial service providers and enterprises from fraudulent activity. Despite the effectiveness of anomaly detection algorithms, they face challenges such as dealing with unbalanced or unlabelled datasets, high-dimensional and correlated data, and the diverse and rare nature of anomalies, making it difficult to accurately distinguish anomalies from normal patterns.

Quantum Machine Learning (QML), which combines machine learning and quantum computing, offers solutions to these challenges by utilizing quantum algorithms to compute complex kernels that are difficult to handle classically \cite{havlivcek2019supervised}. Research into the potential use of quantum algorithms for anomaly detection has shown promising results. For instance, Kyriienko et al.
\cite{kyriienko2022unsupervised} observe a significant 20\% improvement in average accuracy through integrating quantum kernels into one-class support vector machine models, demonstrating the potential effectiveness of quantum methods in enhancing anomaly detection. However, scalability challenges emerged, affecting training and testing times as the data size increased.

In response to this, our previous work \cite{kölle2024efficient} engages in the search for effective and accurate quantum anomaly detection using one-class SVMs by employing two promising approaches: randomized measurements and variable subsampling. Unmitigated randomized measurements offer superior average precision, but their instability and higher time complexity to number of qubits hinder practical application. Methods utilizing variable subsampling approaches that integrate inversion test kernels demonstrate notable improvements in computational efficiency and a reduction in variance. 

This paper introduces two new methods that bring together randomized measurements, variable subsampling and rotated feature bagging to bridge the gap between these two approaches.
This research work addresses the following questions:
\begin{itemize}
	\item Can Variable Subsampling Quantum OC-SVM ensembles benefit from the utilization of Randomized Kernel Measurements and is there a performance trade-off?
	\item Can the use of Rotated Feature Bagging on these Variable Subsampling ensembles assist in reducing the exponential time complexity with respect to the number of qubits/features? 
\end{itemize}
The goal is developing dependable quantum anomaly detection models that achieve a balance between performance, stability, and computational efficiency. Our new methods yield vastly improved average precision and even better time complexity, despite their higher variance.

This work is structured as follows: in \cref{sec:preliminaries}, we give a short introduction to One-Class Support Vector Machines and Quantum Kernel Embeddings. Then, we highlight related studies in \cref{sec:related-work} and outline our approach using Randomized Measurements and Variable Subsampling in \cref{sec:approach}. In \cref{sec:exp}, we provide information about our experimental setup and in \cref{sec:results} we share the results of our study. We conclude in \cref{sec:conclusion} with a short summary and suggestions for further research. 

%% file: content/2_preliminaries.tex
\section{Preliminaries} \label{sec:preliminaries}
In this section, we provide the foundational concepts and methodologies pertinent to our study. We begin with an exploration of One-Class Support Vector Machines, a key technique for unsupervised anomaly detection. Following this, we delve into Quantum Kernel Embedding, illustrating how quantum circuits can be leveraged to encode data into high-dimensional feature spaces for enhanced separability.

\subsection{One-Class Support Vector Machines}
\begin{figure}[tb]
    \centering
        \begin{tikzpicture}[
        declare function={c(\x)=-\x+5;},
        declare function={a(\x)=-sqrt(0.4-(\x)^2);},
        declare function={b(\x)=sqrt(0.4-(\x)^2);},
        declare function={f(\x) = 2*sqrt(2)*rad(atan(\x/(2*sqrt(2))))*5/2.99;}]
        \begin{axis}[width=0.5*\linewidth,
            xmin=-2,
            xmax=2,
            ymin=-1.26,
            ymax=2,
            axis lines=middle,
            axis equal image,
            xtick=\empty, ytick=\empty,
            enlargelimits=true,
            clip mode=individual, clip=false,
            xlabel={\small$d_1$},
            ylabel={\small$d_2$},
            xlabel style={at={(ticklabel* cs:1)},anchor=north},
            ylabel style={at={(ticklabel* cs:1)},anchor=south}
        ]
        \addplot[only marks, mark={x}, samples=120]
            (f(x)+1, {0.5*(a(x)+b(x)) + rand * ( a(f(x)) - b(f(x))) / 2}+1.5);
        \addplot[only marks, mark={x}, samples=120]
            (-f(x)-1, -{0.5*(a(x)+b(x)) + rand * ( a(f(x)) - b(f(x))) / 2}-1.5);
        \addplot[red, thick, only marks, mark={+}, samples=4] (rand+rand, rand+rand);
        \node[below] at (current bounding box.south){\small Input space};
        \end{axis}
        \begin{axis}[width=0.5*\linewidth, 
        at={(0.6*\linewidth,1)},
        domain=-3:5,
        ymin=-5.25,
        axis lines=middle,
        axis equal image,
        xtick=\empty, 
        ytick=\empty,
        clip mode=individual, 
        clip=false,
        xlabel={\small$d'_1$},
        ylabel={\small$d'_2$},
        xlabel style={at={(ticklabel* cs:1)},anchor=north west},
        ylabel style={at={(ticklabel* cs:1)},anchor=south west}]
        \addplot[only marks, mark={x}, samples=10,domain=0.5:5](\x+rnd,{2*rnd-\x+5.5});
        \node[red, thick, inner sep=-2pt] at (-2.65,-0.1) (A1) {\tiny$+$};
        \node[red, thick, inner sep=-2pt] at (-2.5,2.75) (A2) {\tiny$+$};
        \draw[gray] (A1) -- ($(5,0)!(A1)!(0,5)$) node[midway,sloped,below,rotate=360, scale=0.8]{\contour{white}{\scriptsize$b-\mathbf{w}\cdot \Phi(x_i)$}};
        \draw[gray] (A2) -- ($(5,0)!(A2)!(0,5)$) node[near start,sloped,below,rotate=360]{};
        \draw[gray, <->, thick] (0,0,0) -- ($(5,0)!(0,0,0)!(0,5)$) node[midway,sloped,below,rotate=360]{\contour{white}{\scriptsize $b$}};
        \addplot[thick,domain=-3:8]{c(x)} node[below,very near start,sloped,rotate=360, scale=0.85]{\contour{white}{\scriptsize Separating hyperplane}} node[gray, below, very near end, sloped, scale=0.85] {\contour{white}{\scriptsize Anomalous}} node[gray, above, very near end, sloped, scale=0.85] {\contour{white}{\scriptsize Normal}};
        \node[below] at ($(current bounding box.south) + (-3,-5.8)$) {\small Feature space};
        \end{axis}
        \draw[->, gray](0.45*\linewidth,1.5) -- (0.55*\linewidth,1.5) node[gray, midway, align=center,text width=2.33cm]{\scriptsize Feature Map\\ $\Phi$};
    \end{tikzpicture}
    \caption{The linearly inseparable points on the input space are mapped using a quantum feature map $\Phi$ into a feature space where they are linearly separable \cite{aggarwal2017introduction}.}
    \label{fig:margin-hyperplane}
\end{figure}
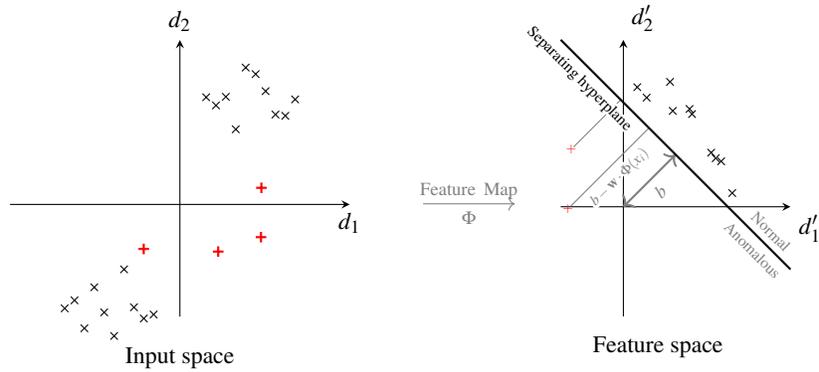

One-class Support Vector Machines (OC-SVM) are popular models for unsupervised anomaly detection proposed by Schölkopf \textit{et al.} \cite{scholkopf1999support}. While conventional SVMs find the maximum margin hyperplane to distinguish anomalies in the labeled data for two or more classes, OC-SVMs assume that the origin represents the anomalous class when labels are absent. In an OC-SVM, the model is trained to distinguish genuine data from anomalies by maximizing the margin $b$ between the origin and the input data, rewarding the model for increasing this margin, and penalizing the points below the hyperplane (\cref{fig:margin-hyperplane}). A hyperparameter $\nu \in (0, 1]$ is introduced, which regulates the ratio of points that remain on the negative side of the separator, which are then subsequently classified as anomalies.

\begin{align}
    \min_{\mathbf{w}, b}\frac{1}{2} \mathbf{w}^2 + \frac 1 {\nu N} \sum_{i=1}^N \max\{b-\mathbf{w}\cdot \Phi(x_i),0\} - b \label{eq:objective-function}
\end{align}

However, real-world data is often not linearly separable, presenting a challenge in identifying a maximum margin hyperplane. Feature maps \(\Phi:\mathcal X \rightarrow \mathcal F\) address this problem by transforming the low-dimensional input data \(\mathcal X\) into a higher-dimensional feature space 
\(\mathcal F\) for improved separability. 
Due to the computational complexity of the direct computation of feature maps, kernel functions  \(k: \mathcal X \times \mathcal{X} \rightarrow \mathbb{R}\) are an alternative approach. These functions determine the similarity of data points within the embedded feature space by using inner products computed directly in the original input space \(\mathcal X\), effectively and efficiently overcoming the need for explicit coordinate computations in the high-dimensional feature space  \(\mathcal F\). 

A cornerstone of kernel-based learning algorithms is the efficiency of the computation of kernel functions  \(k(x_i, x_j\)), which can be concatenated into a Gram matrix $G$:
\begin{align}
   G_{ij} = \langle \Phi(x_i), \Phi(x_j) \rangle = k(x_i, x_j)
\label{eq:gram-matrix-entries}
\end{align}
The Gram matrix permits formulating a dual form of the problem in \cref{eq:objective-function} and enables finding an implicit parameterization of the separating hyperplane during training, determined by the optimal support vectors, which are a subset of the training data identified by the dual coefficients \(\alpha_i\).

The scoring of new data is a function of implicit parametrization and its relation to the separating hyperplane. 
\begin{align}
    \text{Score}(x_\text{new}) = \sum_{i=1}^N \alpha_i \cdot k(x_{\text{new}}, x_i) \label{eq:decision-function}
\end{align}

The assigned label for new data is specified by the score's sign, with negative scores indicating anomalies and positive scores signifying normalcy.


\subsection{Quantum Kernel Embedding}
Unlike classical feature maps, which transform data into a higher-dimensional feature space, quantum feature maps leverage parameterized quantum circuits to directly encode data as quantum states within the Hilbert space $\mathcal{H}$. This encoding is achieved through the application of a data-dependent unitary quantum gate $U_\Phi(x)$ acting on a quantum basis state mathematically expressed as $\ket{\Phi (x)} = U_\Phi(x) \ket{0}$. 
The IQP-like (Instantaneous Quantum Polynomial) feature map, which has the advantage of being hard to simulate classically \cite{havlivcek2019supervised}, encodes a \(d\)-dimensional input \(x_i\) into \(d\) qubits.
 This process starts with applying a block of Hadamard gates, which prepares the qubits in a superposition state. Each input data feature is encoded into individual qubits by single qubit $Z$-rotations, ensuring that the information is accurately represented in the quantum state. $R_{ZZ}$ gates are applied between adjacent qubits to encode the interactions between the features, effectively linking the qubits and embedding the correlations into the data.

\begin{equation}
    \begin{split}
    \ket{\Phi(x_i)} &= U_Z(x_i)H^{\otimes d}U_Z(x_i)H^{\otimes d} \ket{0^d},\\
    U_Z(x_i) &= \exp\left(\sum_{j=1}^{d} \lambda x_{ij}Z_j + \sum_{j=1}^d \sum_{j'=1}^d \lambda^2 x_{ij}x_{ij'} Z_j Z_{j'}\right),
    \end{split}
\end{equation}

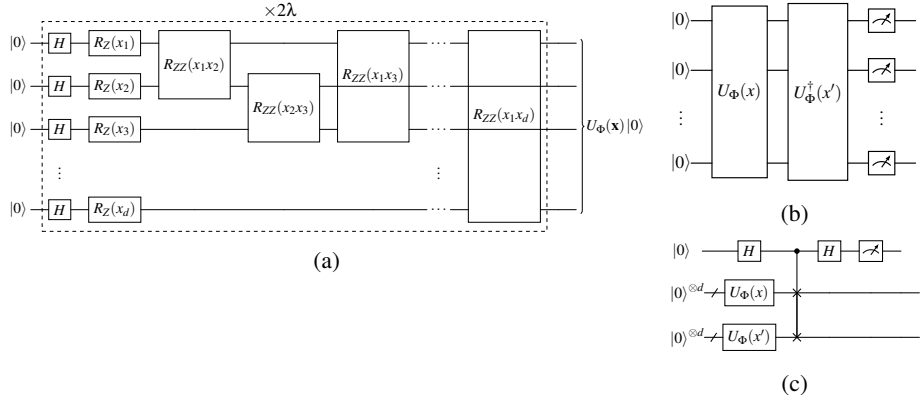
\begin{figure}
\centering

\tabskip=0pt
\valign{#\cr
  \hbox{%
    \begin{subfigure}[b]{.70\textwidth}
    \centering
        \resizebox{\linewidth}{!}{%
        \begin{quantikz}[transparent,font=\large]
        \ket{0} \ & \gate{H}\gategroup[5,steps=7,style={dashed,inner xsep=2pt}, label style={label position=above left}]{{\quad\quad\quad\quad\quad\quad\quad\quad\quad\quad\quad\quad\Large$\times 2\lambda$}} & \gate{R_Z(x_1)} & \gate[2]{R_{ZZ}(x_1 x_2)} &  & \gate[3,label style={yshift=0.3cm}]{R_{ZZ}(x_1 x_3)} &  \ \ldots\ & \gate[5,label style={yshift=0.3cm}]{R_{ZZ}(x_1 x_d)} &  & \rstick[5]{$U_\Phi(\mathbf{x})\ket{0}$}\\
        \ket{0} \ & \gate{H} & \gate{R_Z(x_2)} &  & \gate[2]{R_{ZZ}(x_2 x_3)} & \linethrough & \ \ldots\ & \linethrough &  & \\
        \ket{0} \ & \gate{H} & \gate{R_Z(x_3)} &  &  &  & \ \ldots\ & \linethrough &  & \\
        & \setwiretype{n}\vdots & & & & & \vdots & & & \\
        \ket{0} \ & \gate{H} & \gate{R_Z(x_d)} &  &  &  & \ \ldots\ &   &  & 
        \end{quantikz}%
        }
        \caption{}
        \label{fig:IQP-FeatureMap}
    \end{subfigure}%
  }\cr
  \noalign{\hfill}
  \hbox{%
    \begin{subfigure}{.28\textwidth}
    \centering
  \resizebox{\linewidth}{!}{%
                \begin{quantikz}[transparent,font=\large]
                \ket{0} & \gate[4]{U_\Phi(x)} & \gate[4]{U^\dagger_\Phi(x')}&\meter{} &\\
                \ket{0} &  & &\meter{} &\\
                \setwiretype{n}\vdots &  & &\vdots &\\
                \ket{0} &  & &\meter{} &
                \end{quantikz}
          }
              \caption{}
              \label{fig:inversion-test}
    \end{subfigure}%
  }\vfill
  \hbox{%
    \begin{subfigure}{.28\textwidth}
    \centering
       \resizebox{\linewidth}{!}{%
        \begin{quantikz}[transparent,font=\large]
            \ket{0} \ \ \ & \gate{H} & \ctrl{2} & \gate{H} & \meter{} &  \\
            \ket{0}^{\otimes d}& \gate{U_\Phi(x)}\qwbundle{} & \swap{1} &  &  &  &\\
            \ket{0}^{\otimes d}& \gate{U_\Phi(x')}\qwbundle{} & \swap{-1} &  &  &  &
            \end{quantikz}}
            \caption{}
            \label{fig:swap-test}
    \end{subfigure}%
  }\cr
}

\caption{Quantum circuits for IQP-like feature map, inversion test and swap test}

\end{figure}

The parameter \( \lambda \) depends on the number of data reuploads and indicates how often the quantum feature map is applied to encode the classical data. Note that this parameter affects the kernel bandwidth similarly to the kernel bandwidth parameter \( \gamma \) \cite{Shaydulin_2022}. \cref{fig:IQP-FeatureMap} provides a visual representation, illustrating the IQP-like feature map that encodes a \(d\)-dimensional input \(x_i\) into \(d\) qubits and demonstrates the influence of the parameter \(\lambda\).

Once data has been encoded into quantum states, the measure of the similarity between these states is their fidelity. The fidelity can be expressed in terms of the density matrices, which have the form 
\begin{align}
   F(x, x') = \text{Tr}(\rho(x) \rho(x')).
\end{align}
In the case of pure quantum states, the fidelity is calculated as the square of the overlap of the states, which is given by 
\begin{align}
     F(x, x') = \vert \braket{\Phi(x')}{\Phi(x)}\vert^2.
 \end{align}

Therefore, fidelity measurement techniques can be used for the determination of the kernel matrix elements. The swap and inversion tests represent the predominant approaches among the various available methods. The inversion test, shown in detail in \cref{fig:inversion-test}, measures the overlap between the pure quantum states of two data points $x$ and $x'$ by applying the unitary feature map of  $x$ followed by the adjoint of the unitary feature map of $x'$ . This method doesn't require additional qubits, but results in circuits twice as deep as those used in the swap test and only works with unitary feature maps. The computation of the kernel matrix with this method requires $O(n^2)$ kernel evaluations, where $n$ is the size of the data. In contrast, the swap test, depicted in \cref{fig:swap-test} embeds the quantum states for two data points, $x$ and $x'$, in parallel, using a controlled swap gate to extract similarity information into an additional qubit, which is then measured in the computation base. While this works for pure and mixed states, it requires twice the circuit width of the inversion test, requiring devices with many qubits. It relies on the swap trick, which derives the inner product from the tensor product of density matrices $\rho_i$ and $\rho_j$ utilizing a swap gate $\mathbb{S}$, expressed in \cref{eq:swap-trick}:
\begin{align}\label{eq:swap-trick}
     \Tr(\rho_i\rho_j) = \Tr(\mathbb{S}\rho_i \otimes \rho_j).
\end{align}

%% file: content/3_related_work.tex
\section{Related Work}
\label{sec:related-work}

\subsection{Quantum Anomaly Detection}
This work tackles the time complexity issue raised by Kyriienko \textit{et al.} \cite{kyriienko2022unsupervised}, which combines the one-class SVM with a complex computational kernel based on the IQP-like feature map (\cref{fig:IQP-FeatureMap}) to achieve a 20\% improvement in average precision compared to the traditional benchmark. It expands on our prior work \cite{kölle2024efficient}, which explores two strategies to reduce time complexity in relation to data size: Variable Subsampling ensembles with Inversion Test kernels and single models using the Randomized Measurements kernel.

Hybrid quantum-classical models have emerged as a leading approach in anomaly detection. For example, Sakhnenko \textit{et al.} \cite{sakhnenko2022hybrid} has refined the hidden representation of an auto-encoder (AE) by integrating a parameterized quantum circuit (PQC) with its bottleneck. This approach transitions to an unsupervised model after training, replacing the decoder with an isolation forest, and assessing the performance across various datasets and PQC architectures. In parallel, Herr \textit{et al.} \cite{herr2021anomaly} has adapted the classical AnoGAN \cite{schlegl2017unsupervised} by incorporating a Wasserstein GAN, in which the generator is substituted with a hybrid quantum-classical neural network. The network is subsequently trained using a variational algorithm.

In contrast, methods like the QUBO-SVM mentioned in Wang \textit{et al.} \cite{wang2022integrating} transform the standard SVM optimization problem into a quadratic unconstrained binary optimization (QUBO) problem. This enables efficient solving using quantum annealing solvers, preserving the traditional SVM optimization framework while accelerating accurate prediction by effectively identifying kernel functions, thus allowing for practical real-time anomaly detection.

Ray \textit{et al.} \cite{ray2022classical} explores hybrid ensembles that integrate bagging and stacking techniques using a combination of quantum and classical components, each playing a significant role in anomaly detection. The quantum components encompass various variable quantum circuit architectures, kernel-based SVMs, and quantum annealing-based SVMs. In contrast, the classical components consist of logistic regression, graph convolutional neural networks, and light gradient boosting models. 

\subsection{Efficient Gram Matrix Evaluation}

Quantum kernel methods play a crucial role in various quantum machine learning applications, but they face significant computational challenges. To address these difficulties, two approaches have been proposed: (i) \emph{Quantum-Friendly Classical Methods}, which reduce the number of kernel matrix elements that need to be evaluated, and (ii) \emph{Quantum Native Methods}, which aim to minimize the overall number of measurements required and rely on classical post-processing, which is easily parallelizable or vectorizable.

Randomized Measurement kernels, pioneered by Haug \textit{et al.} in 2021 \cite{haug2021large} and combined with hardware-efficient feature maps. They enable faster kernel measurement, but could only approximate Radial Basis Function (RBF) kernels on both synthetic and MNIST data. In contrast, the classical shadow method, proposed by Huang \textit{et al.} in 2020 \cite{Huang_2020}, employs a similar quantum protocol but diverges in classical post-processing. It provides classical state snapshots through the inversion of a quantum channel, often achieving reduced error in predicting the second Rényi Entropy.

Variable subsampling, which was first introduced by Aggarwal and Sathe in 2015 \cite{aggarwal2015theoretical}, and its advanced counterpart, variable subsampling with rotated bagging, present an efficient approach to ensemble training. The methods use different sample sizes and rotational orthogonal axis system projections to improve both computational efficiency and an adaptive ensemble model training strategy. They have been successfully tested with algorithms like the local outlier factor (LOF) models and the k-Nearest Neighbors algorithm.

The Distance-based Importance Sampling and Clustering (DISC) approach, proposed by Hajibabaee \textit{et al.} \cite{hajibabaee2021kernel}, and Block Basis Factorization (BBF), introduced by Wang \textit{et al.} \cite{wang2019block}, are both variations of matrix decomposition-based methods for kernel approximation. DISC constructs approximation matrices using cluster centroids as landmarks and assumes the kernel matrix to be symmetrical. On the other hand, BBF creates a smaller inner similarity matrix by employing randomized spectral value decomposition on cluster samples, delivering superior performance in comparison to the k-means Nyström method.

%% file: content/4_approach.tex
\section{Approach}\label{sec:approach}

Several methods for efficient gram matrix evaluation have been highlighted in the previous review of related work. In this context, we will concentrate on two specific methods that can be utilized for both symmetric training kernel matrices and asymmetric prediction kernel matrices. While the Classical Shadows and Block Basis Factorization techniques fulfill this requirement, we decide to investigate Randomized Measurements and Variable Subsampling methods due to their intuitive conceptual framework.

\subsection{Randomized Measurements Kernel}

This method is practically employed in kernel calculation for classification by Haug \textit{et al.} \cite{haug2021large} and is suggested for the quantum OC-SVM in Kyriienko \textit{et al.} \cite{kyriienko2022unsupervised}. It tries to attain linear time complexity with respect to the data set size for the quantum kernel calculation by avoiding redundant measurements. This is achieved through acquiring measurements of the respective quantum feature maps of each individual data point in multiple random bases and subsequently aggregating them using classical post-processing. The method notably diminishes the required number of measurements, thereby alleviating the overall computational burden.

The concept of fidelity computation is based on treating the swap operator $\mathbb S$ as a quantum twirling channel $\Phi_N^{(2)}$, as described in Elben \textit{et al.} \cite{elben2019statistical}. Quantum twirling channels are frequently used in error correction. For instance, the application of a 2-fold local quantum twirling channel to arbitrary operator $O$ is expressed as

\begin{align}\label{eq:2-fold-quantum-twirl}
    \Phi_N^{(2)}(O) = \overline{(U^{\otimes 2})^\dagger O U^{\otimes 2}},
\end{align}

with \(\overline{\mbox{\dots}\raisebox{2mm}{}}\) denoting the average over the Haar random unitaries \(U = \bigotimes_k^N U_k\) where the unitaries \(U_k\), applied to the \(k \in \{1, \dots, N\}\) qubit, are sampled from a unitary 2-design.
A unitary \(t\)-design is a finite set of unitaries which approximates the properties of probability distributions over the Haar measure for all possible unitaries of degree less than \(t\), ensuring uniform sampling across unitary matrices.


The authors \cite{elben2019statistical} demonstrate that the average of the second-order cross-correlation of the randomized measurements' outcome probabilities $P_U(.)$ can be expressed as the expectation value of an operator $O$, which applies to a twirling channel and two copies of the quantum state $\rho$.
\begin{align}\label{eq:2nd-cross-correl}
    \sum_{s,s'} O_{s,s'} \overline{P_U(s)P_U(s')} = \Tr(\Phi_N^{(2)}(O) \;\rho \otimes \rho).
\end{align}
The similarity between the right side of this equation and the formula for the swap trick suggests the possibility of representing the purity and fidelity of quantum states in terms of the outcome probabilities of randomized measurements. For this, it suffices to determine the coefficients $O_{s,s'}$ so that $\mathbb{S} = \Phi_d^{(2)}(O)$. The derivation of these coefficients is facilitated by the utilization of Weingarten calculus for Haar-random unitaries and Schur-Weyl duality \cite{Roberts_2017} yielding:
\begin{align}
    O_{s,s'} = d^N(-d)^{-H(s,s')},
\end{align}
where $d$ denotes the dimension of the $N $ qu\emph{d}its and $H(s, s')$ is the Hamming distance between the measured strings $s$ and $s'$.

The formula for quantum fidelities is derived using these coefficients in \cref{eq:2-fold-quantum-twirl} and can be directly used to calculate the kernel entries from the measurement outcomes in a classical post-processing:

\begin{align}
    F(\rho_1, \rho_2) = \Tr(\mathbb{S}\; \rho_1 \otimes \rho_2)= \sum_{s,s'} \underbrace{d^N(-d)^{-H(s,s')}}_{O_{s,s'}} \overline{P_U(s)P_U(s')
    }.
\end{align}

\begin{figure*}[htbp]
    \centering
        \begin{tikzpicture} 
        \node(A)[scale=0.8, draw=black,text width=3cm,align=center]{Sample $r$ random Haar unitaries $U_{Haar}$};
        \node(B)[scale=0.8, right=of $(A.east)$]{
        \begin{quantikz}[transparent,font=\large]
        \ket{0} & \gate[3]{U_\Phi(x)} & \gate[3]{U_{Haar}}&\meter{} &  & \rstick[3]{$s_A, P_U(s_A)$} \\
        \setwiretype{n}\vdots & &  & \vdots &  & \\
        \ket{0} &  & &\meter{} &  &\\
        \ket{0} & \gate[3]{U_\Phi(x')} & \gate[3]{U_{Haar}}&\meter{} &  & \rstick[3]{$s'_A, P_U(s'_A)$} \\
        \setwiretype{n}\vdots & &  & \vdots &  &  \\
        \ket{0} &  & &\meter{} &  &
        \end{quantikz}
        }; 
        \node(C)[scale=0.8, draw=black, right=of $(B)$,text width=3cm,align=center, xshift=4.5cm]{Classical Postprocessing using \cref{eq:qRM-kernel}};
        \draw[->] (A.east) -- (B.west);
        \draw[->, sloped] (B.20) -- (C.west);
        \draw[->, sloped] (B.-20) -- (C.west);
        \draw[thick,dashed]($(B.north west)$) rectangle ($(B.south)+(1.7cm,0)$);
        \draw[thick, ->] (B.53) arc (-90:174:4mm) node[scale=0.8,left,text width=5cm,align=center]{\footnotesize Repeat measurement for $r$ Haar random unitaries, each for $n_s$ shots};;
        \end{tikzpicture}
    \tiny\caption{The protocol and the circuit architecture for calculating quantum kernel functions using randomized measurement.\cite{kölle2024efficient}}
    \label{fig:Randomized-Measurements}
\end{figure*}
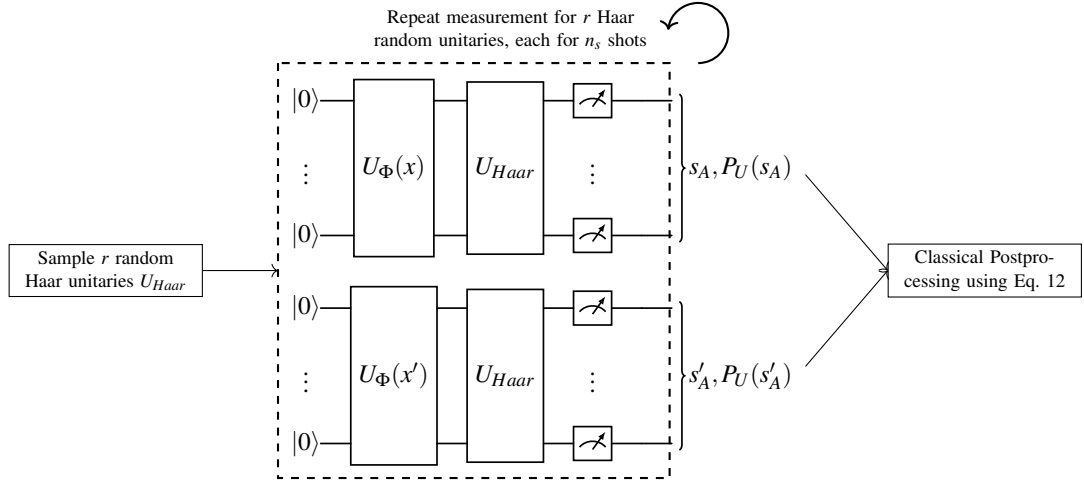

To construct a local Haar random unitary $U_{Haar}$, corresponding to a single basis rotation, a unitary $U_k \in SU(2)$ is sampled for each qubit and the tensor product of the unitaries for all qubits $U_{Haar} = \bigotimes_{k=1}^d U_k$ is built. As shown in \cref{fig:Randomized-Measurements} \cite{kölle2024efficient}, each quantum circuit uses a unitary $U_\Phi$ corresponding to the quantum feature map, along with one of the $r$ different local Haar random unitaries $U_\text{Haar}$. Every circuit necessitates the execution of $n_s$ distinct shots. This process results in the acquisition of $r$ sets of strings, denoted as $s_A$, along with their corresponding measurement probabilities, $P_U^{(i)}(s_A)$. These probabilities are calculated for each string resulting from random basis rotations, represented by $U_\text{Haar}$. Since we are using $N$ qu\emph{b}its, post-processing involves utilizing the following formula:

\begin{equation}\label{eq:qRM-kernel}
    \begin{split}
        K(x_i,x_j) = 2^N \sum_{s_A, s_A'} (-2)^{-H(s_A, s_A')} \overline{P^{(i)}_{U}(s_A)P^{(j)}_{U}(s_A')}
    \end{split}
\end{equation}


The fidelity measurement has a statistical error, which can be approximated as $\Delta G \approx \frac{1}{s\sqrt{r}}$ \cite{haug2021large}. This error needs to be mitigated, especially in noisy hardware situations. One way to do this is by recording purities in the diagonal of the training kernel matrix, computing the purities of new data from their randomized measurements and using them to calculate the mitigated kernel elements:
\begin{align}\label{eq:error-mitigation}
    K_m (x_i,x_j) = \frac{\text{Tr}(\rho_i\rho_j)}{\sqrt{\text{Tr}(\rho_i^2)\text{Tr}(\rho_j^2)}}.
\end{align}

The quantum kernel calculation part uses randomized measurements and has a time complexity of $nrn_s$ ($n$: data size, $r$: basis rotation unitaries, $n_s$: shots per basis rotation). In contrast, classical post-processing requires a time complexity of $n^2$, which increases exponentially as the number of qubits (features) grows. The implementation described in Haug \textit{et al.} can be found in the \texttt{Large Scale QML} in the GitHub repository\footnote[1]{\url{https://github.com/chris-n-self/large-scale-qml}}, and the randomized measurement processing and combination functionalities can be accessed from the repository \texttt{qc\_optim}\footnote[2]{\url{https://github.com/chris-n-self/qc_optim}}.

A novel implementation\footnote[3]{https://github.com/AfraeA/q-anomaly} was developed to accommodate the IQP-like feature map and enable interim kernel copy retention and calculation time-keeping. 

\subsection{Variable Sampling Ensembles using Inversion Test Kernels}
An ensemble technique called Variable Subsampling is recommended in \cite{aggarwal2017introduction} to address sensitivities in the one-class SVM, specifically regarding kernel choice and hyperparameter $\nu$ values. Variable Subsampling ensembles, unlike bagging ensembles, not only select random subsets of the data for model training, but also ensure these subsamples are of different size. This implicitly enables sampling across parameter spaces, particularly those associated with data size, such as the expected anomaly ratio $\nu$ in the one-class SVM.

For instance, if a variable subsampling ensemble consists of 3 OC-SVMs trained with $\nu=0.1$ and sample sizes of 53, 104, and 230, each ensemble component would have a different number of support vectors, leading to different decision boundaries. By combining these, it is possible to mitigate bias or variance in predictions. 

The construction of the ensemble commences by uniformly sampling $c$ different subsample sizes $n_i$ between 50 and 1000. Data subsets $D_i$ corresponding to the sampled sizes $n_i$ are randomly selected from the dataset and utilized to train base model instances, notably the quantum one-class SVM with the quantum Inversion Test. Even though different subsamples might contain the same element, the data within a subsample is sampled without replacement. Predictions are derived by aggregating normalized decision functions from all components. Each ensemble component $i \in \{1, \dots, c\}$ is trained with a different data size $n_i$, resulting in variations of the decision functions \cite{aggarwal2015theoretical}. Averaging outlier scores is advantageous for reducing variance and achieving better performance with smaller datasets. However, utilizing the maximum score decreases bias and increases variance. Following the extraction of decision function values, the class label is determined using the threshold function $\text{sgn}(.)$.

Adding more components and a higher maximum subsample size can lead to better variance reduction, but this comes with the cost of requiring more computational resources and time. Nonetheless, by selecting hyperparameters reasonably, this balance can be effectively managed. For instance, by opting for a maximum subsample size of $n_\text{max} =100$, as opposed to the 1000 points suggested by the original study, and employing \(c = \lfloor \frac {n}{100}\rfloor\) components rather than a fixed count of 100, we might compromise on performance to a certain extent. This approach offers valuable insights into the ensemble method's performance when scalability is a significant consideration.

During the training period, the computational complexity is around $c \times (\frac{n_{\text{min}} + n_{\text{max}}}{2})^2$, with $c$ representing the total number of components in the ensemble, while $n_{\text{min}} = 50$ and $n_{\text{max}} = 100$ denote the smallest and largest subsample sizes respectively. When adjustments for scalability are applied, the complexity formula is adjusted to $\lfloor \frac{n}{100} \rfloor \times \left(\frac{50 + 100}{2}\right)^2$, which shows a linear complexity pattern. The testing phase also follows a linear complexity trend, described by $c \times \frac{n_{\text{min}} + n_{\text{max}}}{2} \times n_{\text{test}}$, where $n_{\text{test}}$ is the number of test samples.


\subsection{Variable Subsampling Ensembles using Randomized Measurements Kernels}
The Variable Subsampling method explained in the section above can also be used in combination with the quantum Randomized Measurements (RM) kernels. This can be done by passing each of the selected subsamples to an OC-SVM instance that uses the quantum Randomized Measurements kernel. 

The aim of incorporating the two methods is that they can potentially reduce the high variance that the unmitigated Randomized Measurements models exhibit in \cite{kölle2024efficient}. This is under the premise that the method combines scores of multiple components, each trained with different samples. 
Similar to Variable Subsampling using the Inversion Test kernel, the method leads to a reduction in time complexity related to data size. However, it has exponential time complexity with respect to the number of features/qubits, comparable to the single OC-SVM model with the randomized measurements kernel.

\subsection{Variable Subsampling Ensembles with Rotated Feature Bagging using Randomized Measurements Kernels}
Variable Subsampling with Rotated Feature Bagging was proposed for classical models along with the standard Variable Subsampling method by \cite{aggarwal2015theoretical}. It makes use of the premise that real data usually contains considerable correlations across different dimensions and can thus be represented in much lower dimensions without causing information loss. 

Since finding the optimal transformation of the data is not trivial, it is much simpler to sample random projections and average the resulting scores from the models trained on the projected data samples.

Therefore, besides reducing the variance across the variously sized samples, the approach reduces the variance across projections of the data onto different lower-dimensional axis systems.

In addition to the steps of the standard variable subsampling method, the data sample $D_i$ of each component $i \in \{1, \dots, c\}$ is projected into a lower-dimensional axis system before it is used in training. This random projection is accomplished by matrix multiplication with a random rotation matrix $E_i$. With an original subsample data matrix $D_i$ of the size $n_i \times d$ and the random projection matrix $E_i$ of the size $d \times r'$, the resulting sample data matrix $D_i' = D_i \cdot E_i$ has size $n_i \times r'$. 

The matrix $D_i'$ is passed as training data for the component $i$. In this case, this component is an instance of a quantum OC-SVM with a Randomized Measurements kernel. The random projection matrix $E_i$ is saved to be eventually reused during testing to project the new data into the same low-dimensional axis system before it is passed to the component $i$ for scoring.

The random projection matrix $E_i$ is unique to each ensemble component. It is calculated by constructing a matrix $Y$ from elements that are sampled from a uniform distribution of values in the range $[-1,1]$ and subsequently using the Gram-Schmidt process to obtain $r'$ mutually orthogonal basis columns  of the matrix $Y$. 

The dimensions of the new axis system are usually set to $r'=2 + \lceil\frac{\sqrt d}{2}\rceil$. This is because of the observation that real data’s implicit dimensionality grows slower than $\frac{\sqrt d}{2}$ in real datasets. The additional two dimensions are added to that number, as the method would otherwise not function for data that comes originally with 3 or fewer dimensions. 

The main purpose of using Rotated Feature Bagging in our case is mainly to address the exponential time complexity to number of features/qubits resulting from the usage of the Randomized Measurements kernel without losing performance. This is a natural result of the reduction of the number of features/Qubits to $r' = 2 + \lceil\frac{\sqrt d}{2}\rceil$. For instance, if the original dataset has 28 features, the components of the ensemble model will be trained with 5 features.

%% file: content/5_experimental_setup.tex
\section{Experimental Setup}
\label{sec:exp}


This section outlines the implementation details to ensure the reproducibility of the experiments. It discusses in detail the preliminary data preparation procedures, the different approaches of kernel computation, and the methods used to select unique hyperparameters.

\begin{enumerate}
    \item 
    The initial set aims to analyze the performance, training and evaluation times related to the size of the dataset. Its primary goal is to understand how model performance evolves with increased amounts of data and to compare the computational effeciency provided by our approaches, as outlined in \cref{sec:approach}.
    \item In our second set of experiments, our goal is to investigate the impact of feature (or qubit) numbers on the performance and  computation times of our approaches. We aim to ascertain whether there are any adverse effects on performance and to showcase the time complexity in function of the number of features/qubits. 
\end{enumerate}
All experiments have been performed with a variety of 15 seeds, ranging from 0 through 14.

\subsection{Datasets}

In our research, we analyze two distinct datasets, categorized based on their origin as either synthetic or real-world, to examine different methodologies. The synthetic dataset is only employed in the first set of experiments since its two-dimensional nature inherently restricts the exploration of a wide range of features.

\subsubsection{The Synthetic Data} is a two-dimensional, not linearly separable dataset developed through alterations made to a demonstration of OC-SVM from SKlearn\footnote{\url{https://scikit-learn.org/stable/auto_examples/svm/plot_oneclass.html}} to generate training samples of various sizes. The test samples for the synthetic data consistently consist of 125 points, featuring an anomaly proportion 0.3.

\subsubsection{The Credit Card Fraud Data} \footnote{\url{https://www.kaggle.com/datasets/mlg-ulb/creditcardfraud}} contains around 284,000 data points, with 492 classified as anomalous (class 1) across 31 features, 28 of which are anonymized features obtained through the application of Principal Component Analysis (PCA). We ommit the 'time' and 'amount' features and use the data in a non-time series manner. The test sets for this dataset also contain 125 points, with an 
anomaly ratio of 0.05.

\subsection{Data Pre-processing}

The quantum kernel measurement technique used and the choice of the dataset determine the  pre-processing methods. The pre-processing of synthetic data occurs only with a randomized measurement kernel. In contrast, the treatment of real data varies based on the quantum kernel measurement technique employed.

\paragraph{Radial Basis Function (RBF) Kernel:} 
When employing the RBF, standard scaling was applied after splitting the data into training and test sets to ensure that all features had a mean of zero and a standard deviation of one. Following this, PCA was employed to reduce the data to the necessary number of features.

\paragraph{Inversion Test Kernels:} The inversion test necessitates an additional procedural step when compared to the methodologies applied to the RBF. This adjustment arises due to the unique application of the data as rotation angles within the construct of the quantum circuit. Consequently, it becomes imperative to implement a scaling factor of 0.1 subsequent to the PCA to accommodate this distinctive requirement.

\paragraph{Randomized Measurements Kernels:}
According to the protocols outlined by Haug \textit{et al.} \cite{haug2021large}, a unique rescaling approach is required for the randomized measurement kernels. After the PCA, a secondary standard scaling phase is carried out, followed by an additional rescaling using the factor $\frac{1}{\sqrt{M}}$, where $M$ represents the post-PCA data dimensionality. 

\subsection{Baselines}

In the initial series of experiments conducted, an attempt is made to replicate the findings associated with the OC-SVM, as outlined in the study by Kyriienko \textit{et al.} \cite{kyriienko2022unsupervised} using the Credit Card Fraud dataset. These findings are subsequently employed as benchmarks for both quantum and classical analyses. 
Due to the need for more detailed information regarding their sampling methodology and the absence of explicit information on the sizes of their test sets, we opted to utilize uniform random sampling to generate our data sets. These sets consist of 500 points for training and 125 points for testing. Consistent with the original authors' approach, we trained the OC-SVM exclusively on non-anomalous data. However, the test set is selected to have a 0.05 ratio of anomalies.

Every seed is associated with a distinct partitioning of the dataset. In the experiments investigating the impact of data, the number of features remains fixed at $d=2$ for the synthetic data and $d=6$ for the CC Fraud dataset, while the training data size varies. For the experiments involving various feature numbers, the data size remains constant at $n = 500$ data samples and the qubit/feature number changes. 

\subsection{Models and Parameter Selection}

In this work, both classical and quantum adaptations of the OC-SVM were utilized, specifically employing the \texttt{OneClassSVM} from the \texttt{SKlearn}\footnote{\url{https://scikit-learn.org/stable/modules/generated/sklearn.svm.OneClassSVM.html}} library. For all classical approaches, the RBF kernel was employed with $\gamma = \frac{1}{N\cdot \text{Var}(M)}$ and alongside a constant $\nu$ value of $0.1$. 

Quantum circuits are crucial for kernel calculations and are created using the \texttt{qiskit} library. They are simulated through \texttt{qiskit\_aer.QasmSimulator}. A $\lambda = 3$ data reuploading was used for all quantum feature maps, and inversion test kernel elements were determined using 1000 shots each.

Randomized measurement circuits were realized with $r=30$ measurement settings and subjected to $s=9000$ shots each. They require efficient classical post-processing, including minimal embedded loops and prioritizing efficient matrix operations. The variable subsampling method, utilizing $c=\lfloor\frac{n}{100}\rfloor$ components ($n$ representing the training set size) and a subsample size $n_i \in [50, 100]$ to ensure scalable model performance, invokes the Inversion test or the Randomized Measurements kernel using the same parameters mentioned above for the single OC-SVMs. A consistent value of $\nu$ is employed in the calculation of the desired kernel matrix of each ensemble component. 

\subsection{Performance Metrics for Imbalanced Datasets}\label{subsec:PerformanceMetrics}
In the context of evaluating the performance of anomaly detection models on data sets that are primarily imbalanced, it is essential to consider factors beyond the conventional measure of accuracy. This is because accuracy may not be the most comprehensive measure of a model's performance across diverse categories of data. As Aggarwal \cite{aggarwal2017introduction} highlights, metrics such as the F1 score and average precision are much more appropriate for this type of data. These metrics, which consider precision and recall, provide a more accurate assessment of a model's ability to identify anomalies within a highly imbalanced data set.

\paragraph{F1 score} is the harmonic mean of precision and recall, providing a balanced perspective on model performance with regard to false positives and false negatives. It is defined as follows:
\begin{align}
    \text{F1 score} = 2\cdot \frac{\text{Precision}\cdot\text{Recall}}{\text{Precision}+\text{Recall}}.
\end{align}

\paragraph{Average Precision} meticulously quantifies a model's proficiency in anomaly detection across a spectrum of thresholds. This is achieved by calculating the area encompassed beneath the precision-recall curve, which provides a detailed assessment of the model's capacity to discriminate anomalous occurrences.

\begin{align}
    \text{AP} = \sum_k [\text{Recall}(k) - \text{Recall}(k+1)] \cdot \text{Precision}(k).
\end{align}

\noindent In terms of average precision, a ratio equal to the data's anomaly ratio indicates a model with no anomaly detection capability. Conversely, a score of 1 signifies a flawless detector. Accordingly, the focal point of our analysis encloses average precision, augmented by an examination of supplementary metrics, namely precision, recall, and the F1 score, to provide a comprehensive evaluation.

%% file: content/6_results.tex
\section{Results}
\label{sec:results}
In the following section, we evaluate the different methods to lower the time complexity of our model while maintaining performance. We focus on the average precision, which allows us to assess the performance without taking into consideration the threshold used on the scores, but also reports the evolution of thresholded metrics like the F1 score, the recall, and the precision.
\begin{figure*}[htbp]
    \centering
    \includegraphics[width=\linewidth]{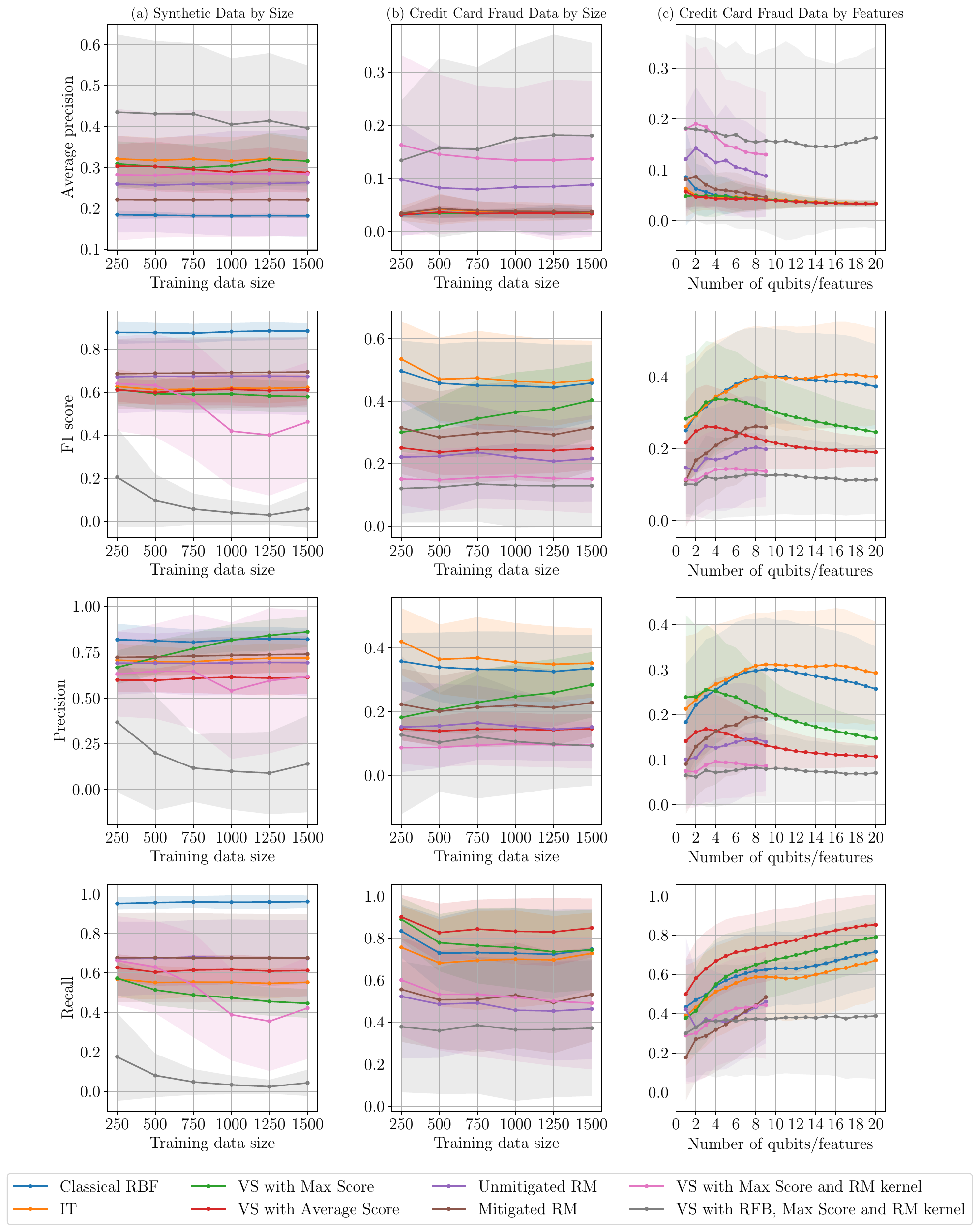}
    \caption{The mean and standard deviation bars for the performance of the OC-SVM models trained using the methods presented. Column (a) represents the results for the synthetic data (2 features). Column (b) shows the results for the Credit Card Fraud data (6 features). Column (c) corresponds to the experiments with 500 points and varying feature numbers.}
    \label{fig:performance}
\end{figure*}
\subsection{Performance Relative to Data Size}

In \cref{fig:performance} columns (a, b) show the performance metrics for the synthetic dataset and the Credit Card Fraud Dataset in relationship to the data size, respectively.
\paragraph{The Quantum Inversion Test (IT) OC-SVMs } appear to reach higher average precision than the classical RBF model for the synthetic dataset and similar average precision to the classical RBF on the Credit Card dataset. Since the average precision of a random detector is equal to the anomaly ratio, which is 0.3 for the synthetic dataset and 0.05 for the Credit Card Fraud dataset, it seems like the method performs slightly better than a random detector on the synthetic dataset and similar to one on the Credit Card Fraud dataset. The plots further showcase that models trained with this kernel have higher variance then the classical RBF.

\paragraph{The Quantum Randomized Measurements (RM) OC-SVMs } surpass the classical RBF but not the single Inversion Test models in average precision for the synthetic dataset.
On the Credit Card Fraud dataset, the unmitigated version outperforms all the previously evaluated methods on average precision and is only topped by the ensembles that employ the quantum Randomized Measurements kernel. 
The models with the unmitigated kernels showcase higher average precision but are more unstable because of their higher variance.

\paragraph{The Variable Subsampling ensembles using the quantum Inversion Test kernel (VS-IT)} results approximate those of the quantum Inversion Test for both dataset. The variation with the maximum score seems better for the synthetic dataset.

\paragraph{The Variable Subsampling ensembles using the quantum Randomized Measurements kernel (VS-RM)} exhibit a downward slope for the average precision until 750 samples, where the trend appears to stabilize. The variance for this method is high, similar to the single models trained on the randomized measurements kernel, but decreases with an increasing number of features. 

\paragraph{The Variable Subsampling with Rotated Feature Bagging ensembles using the Randomized Measurements kernel (VS-RFB-RM)} obtain better average precision and do not experience the same downward trend with increasing data size, demonstrating the potential benefit of using different rotated axis systems in uncovering anomalies.

\subsection{Performance Relative to Qubit Number / Feature Number}

\paragraph{The Quantum Inversion Test OC-SVMs (IT)} achieve a negligible improvement in average precision compared to the classical radial basis function (RBF) models. This contradicts the results of Kyriienko et al. \cite{kyriienko2022unsupervised}. This discrepancy in the results might stem from differences in the number of runs, the seeds used for the experiments, and the data selection method, as these details were not disclosed in their paper.

\paragraph{The Quantum Randomized Measurements OC-SVMs (RM)} performance seems to decline when increasing the qubits. When error mitigation (\cref{eq:error-mitigation}) is applied during the gram matrix calculation, the results are close to the classical RBF and the quantum inversion test models. But if the error mitigation is skipped, the average precision is higher than all the previous methods. The variance of the method noticeably exeeds the previous methods, especially if the unmitigated version is selected, and does not reduce as fast when increasing the number of features/qubits.

\paragraph{The Variable Subsampling ensembles using the quantum Inversion Test kernel (VS-IT)} achieve similar results to the single quantum Inversion Test models and have slightly lower variance than the single models. The difference in variance decreases with increasing number of features/qubits. 

\paragraph{The Variable Subsampling ensembles using the quantum Randomized Measurements kernel (VS-RM)} achieves better average precision than the previous methods and is only surpassed by its variant with rotated feature bagging. The variation is very high, even in comparison to the single models with Randomized measurements kernels. However, we notice a slight improvement in the variation with higher qubits/features numbers.

\paragraph{The Variable Subsampling with Rotated Feature Bagging ensembles using the Randomized Measurements kernel (VS-RFB-RM)} outperforms all the other methods in average precision with increasing number of features/qubit. While it might seem that there is little gain from using 7 principal components (from PCA) instead of 5, since the components will transform the data to 3-dimensional data in both cases, the transformation uses all the dimensions passed to it. It is clear that the method decreases the dependence of performance on the number of features, and couples it with the number of ensemble components instead, as that allows using the information from the features rotations better. Thus, there is potential for further improving the results of the method, if the number of the ensemble OC-SVM components is increased and more data is used.

\begin{figure*}[htbp]
    \centering
    \includegraphics[width=\linewidth]{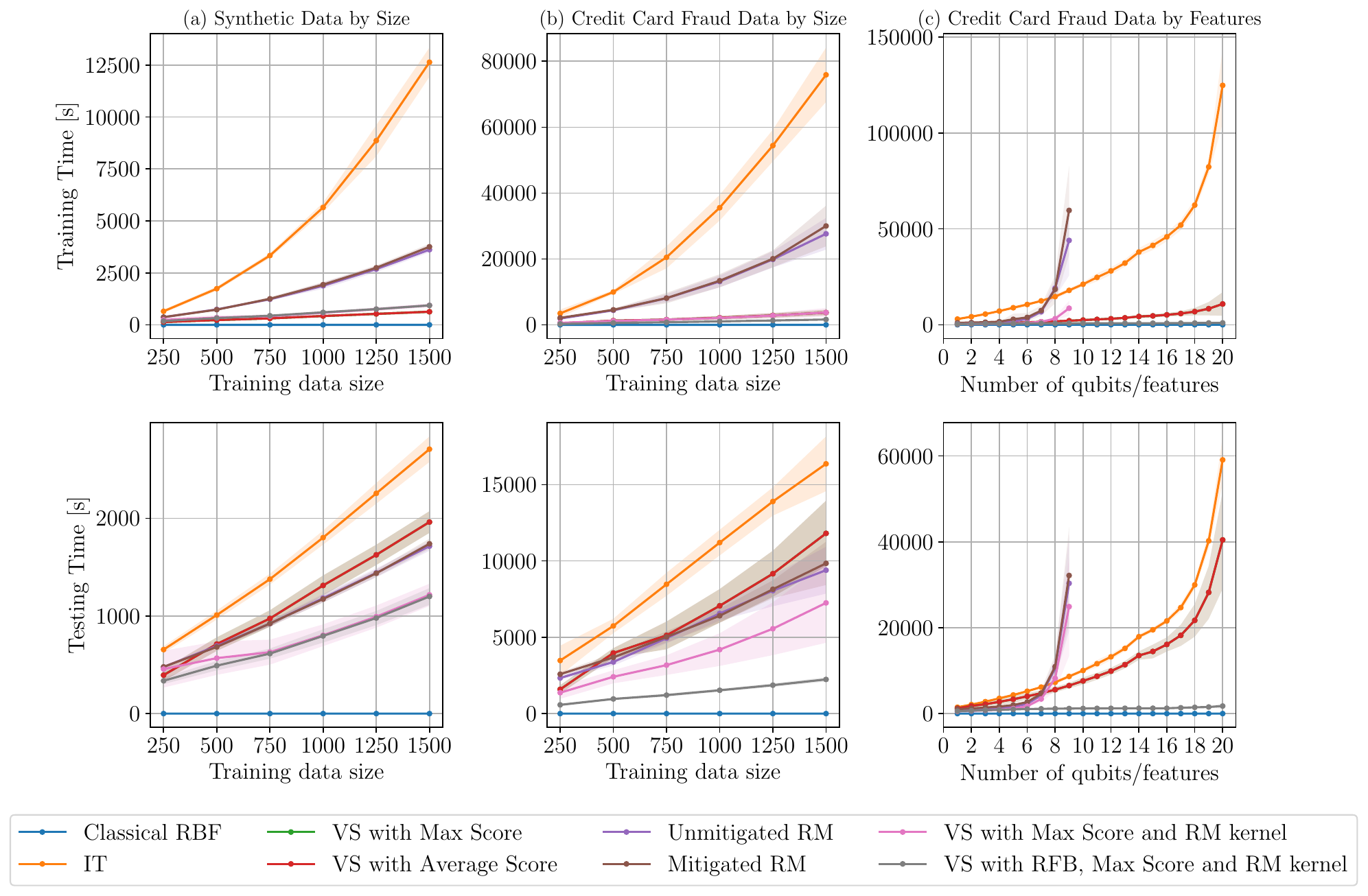}
    \caption{Training and testing times in seconds, based on the data size in columns (a) and (b), and on feature/qubit numbers in column (c).}
    \label{fig:times}
\end{figure*}

\subsection{Time Complexity Relative to Data Size}
\cref{fig:times} (a - b) display the training and testing times for our models in seconds in relationship to the data size for the synthetic dataset and the Credit Card dataset. We obtain consistent results for both datasets.

\paragraph{The Quantum Inversion Test OC-SVMs (IT)} evolves quadratically with the training data size during training and linearly during testing.
\paragraph{The Quantum Randomized Measurements OC-SVMs (RM)} substantially shortens the training times, attaining an approximate 50\% reduction in training time whilst tripling the data size. However, the time complexity based on data size remains quadratic. It is worth noting that the error mitigation in \cref{eq:error-mitigation} does not seem to significantly increase the kernel calculation time.
\paragraph{The Variable Subsampling ensembles using the quantum Inversion Test kernel (VS)} training times coincide with those of the other variants Variable Subsampling methods. The method extensively reduces the training times in comparison to the usage of a single Inversion Test OC-SVM, and successfully achieves linear time complexity to data size. 
\paragraph{The Variable Subsampling ensembles using the quantum Randomized Measurements kernel (VS-RM)} achieves linear complexity to data size and even higher reductions in time, in comparison to the VS with the Iversion Test kernel.
\paragraph{The Variable Subsampling with Rotated Feature Bagging ensembles using the Randomized Measurements kernel (VS-RFB-RM)} have training times which are similar to the other variable subsampling variants, but are especially more effective in diminishing testing times. Notably, we obtain a nearly 90\% reduction in training times and an 80\% reduction in testing times when using triple the amount of training data. For this method, the time complexity to data size for this method during both training and testing is linear.

\subsection{Time Complexity Relative to Qubit Number / Feature Number}
\cref{fig:times}(c) exhibits the training and testing times in function of the number of features/qubits of our models for the Credit Card dataset.
\paragraph{The Quantum Inversion Test OC-SVMs' (IT)} time complexity during training seems to evolve quadratically with the number of qubits/features. 
\paragraph{The Quantum Randomized Measurements OC-SVMs (RM),} in accordance with the results in \cite{haug2021large}, appear to have exponential time complexity in function of the number of qubits/features. This causes the method to become unusable for high-dimensional datasets without prior usage of dimensionality reduction techniques like PCA.
\paragraph{The Variable Subsampling ensembles using the quantum Inversion Test kernel (VS)} display a quadratic time complexity in relationship to the number of qubits/features, due to the usage of the Inversion Test. The method leads to a drastic reduction in training times, although the the effect on testing times is less pronounced.
\paragraph{The Variable Subsampling ensembles using the quantum Randomized Measurements kernel (VS-RM)} deminish training times more considerably than testing times. The time complexity of this approach has an exponential relationship to the number of qubits/features, similar to the Randomized Measurements kernel method.
\paragraph{The Variable Subsampling with Rotated Feature Bagging ensembles using the Randomized Measurements kernel (VS-RFB-RM)} successfully manage to mitigate the exponential relationship of time to number of qubits arising from the usage of the Randomized Measurements kernel, resulting in significant reductions of training and testing times. The reason for this reduction is that the ensemble components are trained with $2 + \lceil\frac{\sqrt d}{2}\rceil$ dimensional rotated features instead of the original principal components $d$. The usage of rotated feature bagging allows the application of Randomized measurements kernels in use-cases where the data is high-dimensional. This combined with the variable subsample sizes, produces the fastest method to train and test among all the quantum methods we try. 

%% file: content/7_conclusion.tex
\section{Conclusion}
\label{sec:conclusion}
Our study examines multiple approaches for efficient quantum anomaly detection using one-class SVMs. The first approach is based on the classical Variable Subsampling ensemble method, while the second utilizes the quantum Randomized Measurements kernel calculation method. 
Our results demonstrate that the Variable Subsampling method can effectively be used to train OC-SVM-based ensembles with the quantum Inversion Test kernel in linear time complexity to data size and quadratic complexity to the number of features. The method leads to a drastic acceleration in training and testing times, without compromising the performance of the models, provided that the scoring threshold is adjusted.  Alternatively, the unmitigated Randomized Measurements kernel seems to attain higher average precision than the Inversion Test and RBF kernel-based methods for the Credit Card dataset, even though this dataset exhibits higher dimensionality and imbalance than the synthetic dataset. This method however produces unstable models, evident in the high variance. Furthermore, it comes with exponential time complexity to the number of qubits/features and a quadratic time dependence to data size.
These findings motivate a novel approach that integrates both methods, which we test in the same experimental settings. We create new Variable Subsampling ensembles using OC-SVMs trained with the Randomized Measurements kernel. This method surprisingly leads to an increase in performance along with further improvements in training and evaluation times, but still suffers from high variance and exponential time complexity to the number of qubits.
To overcome these drawbacks, we develop a further method, variable subsampling with rotated feature bagging in combination with the Randomized Measurements kernel OC-SVMs. The resulting models, not only have linear time complexities to both the data size and the number of features/qubits, but also achieve higher average precision than the previous models. However, their variance remains very high, which can be attributed to the usage of the Randomized Measurements kernel, the maximum combination function for the scores, and a potentially too low number of ensemble components and maximum subsample size.
These promising results for the time complexity open the door for more future research directions. These include refining the Variable Subsampling methods by using a higher number of components and maximal subsample sizes, thereby directing the focus toward the methods' performance prospects or trying out the proposed methods with alternative feature maps, including learnable ones.
Importance sampling can be employed to improve the Randomized Measurements kernel in the single OC-SVMs, as well as in the Variable Subsampling ensembles. Furthermore, the classical shadow method, which showcases lower average error compared to randomized measurements, can be assessed as an alternative to the Randomized Measurements method in the models above. Finally, it could be interesting to evaluate the promising DISC and Block Basis Factorization methods against the ones proposed in this work.

%% file: content/8_acknowledgements.tex
\section*{Acknowledgements}
This research is part of the Munich Quantum Valley, which is supported by the Bavarian state government with funds from the Hightech Agenda Bayern Plus.